# Aspect Extraction and Sentiment Classification of Mobile Apps using App-Store Reviews

Sharmistha Dey[1]

[1]Doctoral Candidate, Department of Management Studies,
Indian Institute of Technology Madras, Chennai, India
*ms15d022@smail.iitm.ac.in*

**Abstract:** Understanding of customer sentiment can be useful for product development. On top of that if the priorities for the development order can be known, then development procedure become simpler. This work has tried to address this issue in the mobile app domain. Along with aspect and opinion extraction this work has also categorized the extracted aspects according to their importance. This can help developers to focus their time and energy at the right place.

## 1. Introduction

Finding customer requirement is always a big concern for product and service companies. Surveys, focus groups etc. are few of the most popular approaches used by the industries to identify the voice of customers (VoC). However, with increasing availability of online reviews, many of the establishments are showing interest to mine the web to find the VoC. This requirement of mining very large set of unstructured data to find useful pattern is also fuelling the academic community for advancing research in this field. However, up-till now the bulk of the work have been based on the product industries (ex: mobile, camera, laptops etc) [1,2,3,4], service industries (ex: hotel, travel etc.) [5] and even education [6]. The software industry is still a new field to be explored. The main aim of this project is to identify customer requirements for a sector of app (*mobile applications*) using online reviews and customer satisfaction model like Kano's model [7].

As a first step, different aspects (ex. *chat*, *sms*, *theme etc*.) of an entity [8] (here *messenger apps*) will be identified from a large set of app-store reviews. As a next step these aspects will be bucketized into 5 buckets prescribed by Kano's customer satisfaction model [7]. This step will be carried out by conducting a customer survey to rate each aspect with respect to five buckets.

The primary reason behind bucketization of the aspects is the assumption that not all aspects hold the same importance in the eyes of a user. For example, being able to send and receive messages is a must have quality for a messenger app however availability of profile theme may just be considered as a delighter quality. Identifying these buckets and then sentiment score for the aspects belong to a bucket will help a developer to find and prioritize the aspects to be improved or added (from the sentiment score of competitors' apps).

Third step will calculate sentiment score for each attribute. Lastly the results will be summarized for better understanding.

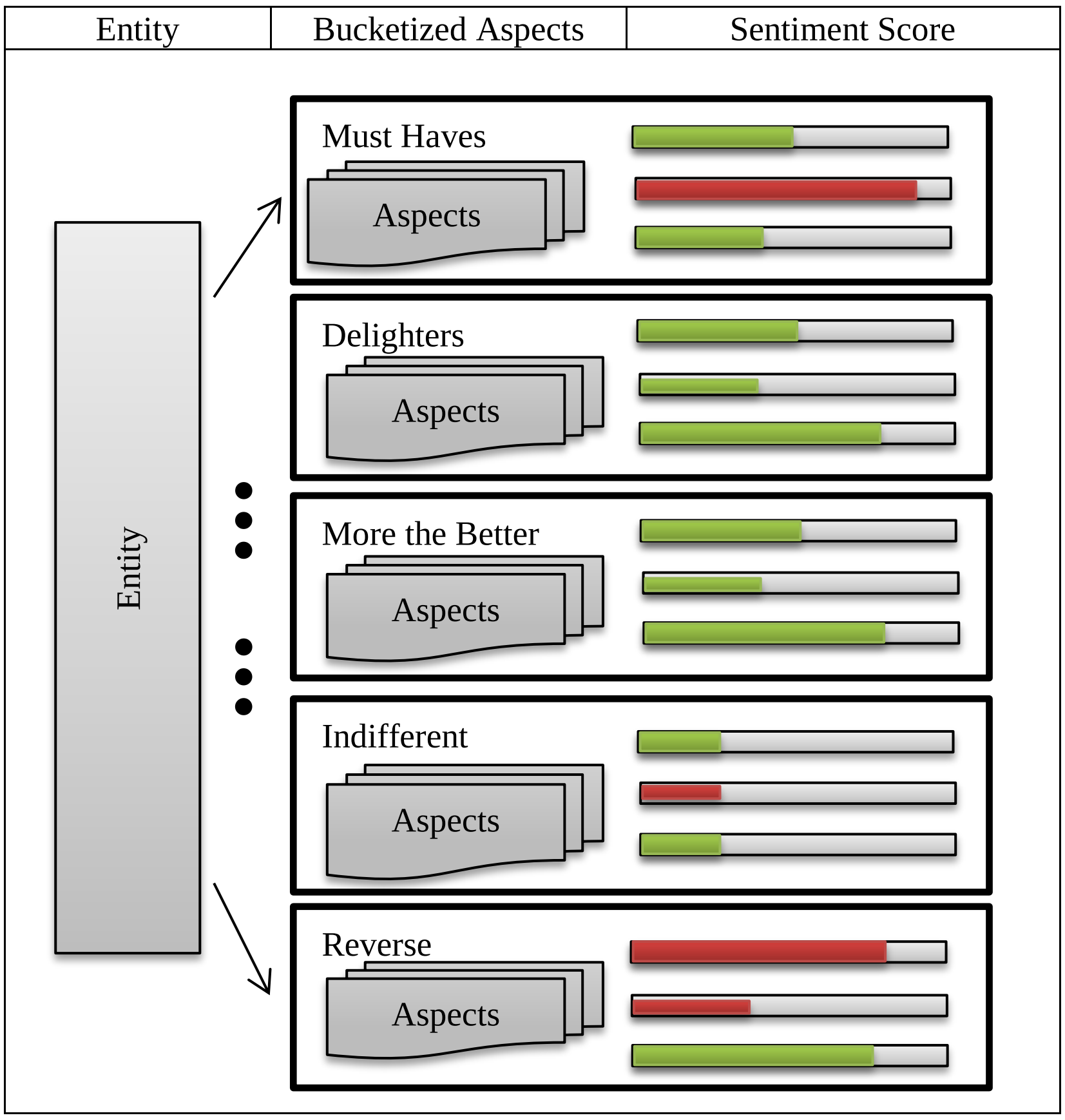

**Fig 1:** A Diagram of the Proposed Work

**Structure of the Paper.** After briefly introducing the aim of the project in the section-1, this paper conducted a quick literature survey in the Section-2. Section-3 and 4 described the dataset and methodologies employed respectively. An evaluation measure is proposed in the Section-5. Section-7 concludes the paper after briefly touching upon the future scopes in section-6. The references used in this paper are listed in Section-8.

## 2. Literature Survey

The job of extracting aspects and sentiments from online reviews of a product is a subcategory of information extraction (IE) task. However the major difference is the structure of the text. The vanilla IE comprises of named entity recognition (NER) and relation extraction from well-structured documents. However, most of the online product reviews are unstructured text. Moreover, mobile app reviews are extremely unstructured, due to the fact that most of the reviews are entered from a handheld device interface. This gives these reviews an essence of chat language resulting into single word entries and large amount of spelling mistakes.

> "*Loveeeeeeeeeeeeeeeeeeeeeeeee it*"

The opinion mining task for a product has three sub-tasks: (1) Aspect and/or entity extraction, (2) Sentiment scoring and (3) Summarizing. The following subsections describe each of these steps briefly.

### 2.1 Aspect and Entity Extraction

In his book [8] Liu defined each opinion as a quintuple $(e, a, s, h, t)$, where $e$ is an entity and $a$ is one of its aspects. $h$ represents the opinionator who expresses her opinion $s$ at timeframe $t$. Names of products, service, individuals, event and organizations are commonly referred as entities. And aspects refer to the components and/or attributes of entities. In this work we are interested in $a$ and $s$, since $e, h$ and $t$ are known.

> "*I think WhatsApp need to update provide like a automatically changing the dp & status as pre set time*"

In the above review sentence *WhatsApp* is the entity, whereas *dp* (display picture) and *status* are the aspects of WhatsApp messenger.

.

There can be two types of aspects: *explicit and implicit*.

> Explicit: "*I need whats app* ***video call feature****.if this feature is provided I will give u five stars*"

> Implicit: "*It is realy simple*"

This work is only going to concentrate on the explicit aspects, which are nouns and noun phrases [8].

As mentioned earlier, aspect extraction is an information extraction task [8]. However in case of aspect opinion analysis the fact that *an opinion always has a target* [8] can be made use of. This means that the opinions are now dependent on the entity or as-pect it was targeted for. And this syntactic relation can be exploited to perform this task. There are four approaches to extract explicit aspects:

1. By finding frequent noun and noun phrases [1]
2. By exploiting syntactic relations
   a. Syntactic dependency: opinion and target relation
   b. Lexical syntactic: by finding pattern in combination of target and opinion word.
3. Using supervised learner
4. Using topic model

Since this work used the frequency based aspect extraction, a short description of that follows.

**Frequency Based Aspect Extraction**: This method can be used only when a large number of reviews are available in the same domain [1]. The first step of this process is to identify the nouns and noun-phrases (NP). This task can be done by part of speech tagging (POS) and NP-chunking. In the next step data mining algorithms are employed to come up with a candidate item set. In [1] association rule mining (ARM) algorithm was used to perform the candidate generation task. Then the candidate space was pruned to get a compact list of aspects.

### 2.2 Aspect Sentiment Classification

Like document or sentence label sentiment classification, aspect sentiment classification also has two approaches: supervised and unsupervised [8].

**Supervised**: In this approach label data marking sentiment orientation of a sentence is already available. The classification task is performed using either SVM (support vector machine) or Naïve Bayes classifiers. The difference of this method from the document or sentence label classifier lies in the fact that the features are now dependent on the entity or aspect.

**Unsupervised:** This is a lexicon based approach. In this approach as well, the extracted features are dependent on the entity or aspect. The same method in supervised learning is used to find the features and then the sentiment orientation is scored using a large set of lexicon. This work made use of the positive and negative lexicons accumulated by Hu and Liu [9].

### 2.3 Kano's Customer Satisfaction Model

Kano's customer satisfaction model [10] categorizes the product or service attributes into following five categories [7].

**Must-have Quality**. These attributes are taken for granted when fulfilled but result in dissatisfaction when not fulfilled.

**More the Better**. These attributes result in satisfaction when fulfilled and dissatisfaction when not fulfilled. These are attributes that are spoken and the ones in which companies compete.

**Delighters**. These attributes provide satisfaction when achieved fully, but do not cause dissatisfaction when not fulfilled. These are attributes that are not normally expected

**Indifferent Quality**. These attributes refer to aspects that are neither good nor bad, and they do not result in either customer satisfaction or customer dissatisfaction.

**Reverse Quality**. These attributes refer to a high degree of achievement resulting in dissatisfaction and to the fact that not all customers are alike

## 3. Description of Data

**Table 1**. Number of Reviews Used

| Kik | Hike | LINE | WhatsApp | SnapChat |
|---|---|---|---|---|
| 3949 | 4742 | 502 | 4868 | 3369 |

Total 17430 reviews from 5 well known messenger apps were used for this project. Among these single and two word entries were not considered for the parsing step. These reviews were crawled from Google App-store between 19th Oct and 22nd Oct 2016. All the 5 apps are in the market for quite a long time and also they have something different to offer in terms of aspects beside the basic offerings.

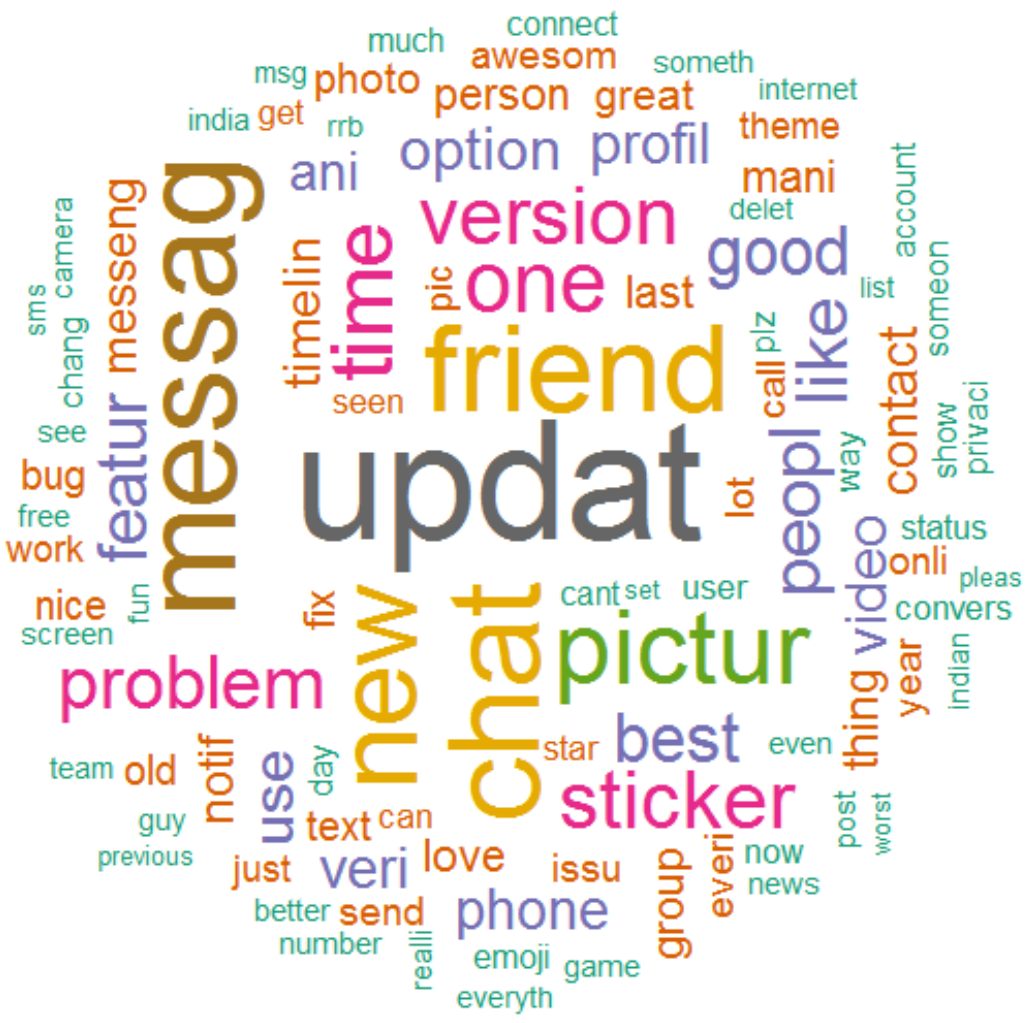

**Fig 2:** Word Cloud containing 100 words ordered using term-frequencies

## 4. Methodology and Analysis of Results

### 4.1. Frequent Feature Extraction

**POS Tagging and Chunking**. Stanford parser [11] was used to perform both POS tagging and probabilistic parsing. The resultant parse trees were used to get the NPs for the further processing.

**ARM.** The NPs extracted in the previous steps were used to generate the transactions required for ARM. The set of transactions had the individual NPs as observations and the unique terms present in the NP as the item set. *Apriori* algorithm of the R package arules [12] was used to find the frequent feature set. 0.04 % support and 60.0 % confidence was used for the rule generation. This setup resulted into 680 rules.

**Pruning**. Next the 680 were pruned to get a compact set of aspects. First each single words ($w_i$) present in the rules were accumulated along with its supersets ($s_{wi}^{j}$) (rules that contained that word). Next it was checked if the difference between number of occurrences of $w_i$ and the number of occurrences of $j^{\text{th}}$ superset $s_{wi}^{j}$ is more than a threshold (*here 3*). If it is so, the single word was considered as an aspect word.

For the multiple word case, it was checked if a sequence of words occurred in the reviews keeping their order intact. Most of the extracted aspect terms were single or two words, with an exception of 'end to end encryption'. Finally, 51 aspect terms were extracted. These terms were then manually combined in case they represented to the same concept. These resulted into 24 categories with 51 aspect words. In the sentiment extraction step the sentiment score of each 51 aspects were calculated and then added up for each category.

### 4.2. Bucketization

The next step bucketized each of 24 aspect categories into the 5 buckets of Kano's customer satisfaction model [7]. This task was completed by performing a survey amongst messenger app users. 31 subjects were chosen for this survey. Table 2 shows their demography. These subjects marked each aspect-category to a bucket they think was most suitable. The majority vote was used for final bucketization.

**Table 2.** Demography of the Survey Subjects

| | |
|---|---|
| **Age** | between 25 and 35 |
| **Education** | minimum Bachelors' degree |
| **Selection Criterion** | Regularly uses different kind of messenger apps |

### 4.3. Aspect Sentiment Classification

The sentiment classification has been performed using unsupervised lexicon method.

The list of review sentences were scanned for an aspect term. If found the sentence was then scanned for positive and negative sentiment words. Next the set of positive or negative sentiment words were scored according to equation (1). The individual sentiment word scores are summed up to get the aspect's positive and negative scores. To identify a word's sentiment orientation Hu, Liu's positive and negative sentiment word lists were used [9].

For each positive sentiment word found it was marked with +1 and for each negative word found it was marked with -1. Then the distance between the aspect word and the sentiment word was calculated and two separate score for positive and negative sentiment was assigned using the equation (1).

$$a_i ss = \sum_j \frac{se_j.ss}{dist(se_j, a_i)} \quad (1)$$

$a_i ss$ is the sentiment score of an aspect $a_i$
$se_j.ss$ is the sentiment score of sentiment expression $se_j$
$dist(se_j, a_i)$ is the distance between $a_i$ and sentiment expression $se_j$

### 4.4 Summarization and Interpretation of Results

Table 3 and Table 4 summarized the results of this project. Table 3 shows the overall positive and negative sentiment score for each aspect. The 51 extracted aspects are first manually put together in 24 categories depending upon their similarities. These 24 categories are then subdivided into 5 buckets (*first and second columns of Table 3 and Table 4*). The positive and negative columns tabulated the aspect sentiment scores calculated using equation (1). The last column shows a sentiment bar showing percentage of positive (green) and negative (red) sentiment score out of total sentiment score (sum of positive and negative).

Table 4, shows the scores for each 5 entities. To make the comparison fair, the original sentiment scores are divided by the total number of reviews used for each entity (since this number varies). The values showed in the table are all in the scale of 10e-4.

From these two tables it can be concluded that the prevalence of positive is higher than the negative. *This may be due to the reason that a messenger application itself is a delighter than a necessity*. However, the most negative feeling is around the updates. A closer look in the reviews showed that many a time the updates cause the app to malfunction therefore results into negative emotions. On the other hand the positive emotion score is comparable with the negative emotion score in case of updates. Some of the aspects did not get any positive or negative score, a closer look of the reviews showed that most of the time reviewers made neutral statements about these aspects. If the must have qualities (chat, text, messages etc.) of the 5 apps are compared (Tablr 4), it can be seen that Hike and Kik gets more positive reviews than WhatsApp. Whereas, Hike and Kik got higher negative score compared to WhatsApp on the update aspect. Hike gets very high positive score for its emoji and stickers. Kik also has a issue with freezing.

Table 3: Overall sentiment score for each aspects, grouped and then bucketized

| Bucket | Aspects | Positive Score | Negetive Score | Colour Bar |
|---|---|---|---|---|
| Must Haves | end to end encryption, privacy | 23.257 | 13.726 | 0.63 |
| | sms, message, text, chat | 664.893 | 224.012 | 0.75 |
| | friends, family | 193.974 | 55.501 | 0.78 |
| | voice, voice call, video call, conf call | 44.304 | 13.638 | 0.76 |
| | video, picture | 181.655 | 168.570 | 0.52 |
| | freeze, restart | 23.315 | 204.577 | 0.10 |
| | group, group chat, admin | 45.642 | 23.017 | 0.66 |
| | content, unknown content | 9.090 | 18.683 | 0.33 |
| Delighters | update, beta, beta tester, newer, older version | 229.424 | 271.856 | 0.46 |
| | timeline | 30.306 | 22.505 | 0.57 |
| | camera | 49.937 | 51.663 | 0.49 |
| | internet connection | 0.000 | 0.000 | |
| | last seen | 0.000 | 0.000 | |
| | sticker, emoji | 293.525 | 68.054 | 0.81 |
| | notification | 23.049 | 33.747 | 0.41 |
| | history, chat history | 2.952 | 2.271 | 0.57 |
| Indifferent | offer, spin | 10.738 | 2.222 | 0.83 |
| | voice changer | 0.000 | 0.000 | |
| | bubble, chat buble | 11.404 | 3.149 | 0.78 |
| | news, cricket | 37.878 | 12.066 | 0.76 |
| | games, teen patti | 29.323 | 5.388 | 0.84 |
| | profile picture, profile, status, theme, home screen | 74.418 | 49.418 | 0.60 |
| | hashtag | 11.411 | 5.756 | 0.66 |
| Reverse | phone number | 0.000 | 0.000 | |

**Table 4:** All values are$(10e-4)$. The sentiment score for individual entities divided by the total number of reviews per entity

| Bucket | Aspects | WhatsApp | | Hike | | Kik | | Line | | Snapchat | |
|---|---|---|---|---|---|---|---|---|---|---|---|
| | | Positive | Negetive | Positive | Negetive | Positive | Negetive | Positive | Negetive | Positive | Negetive |
| Must Haves | end to end encryption, privacy | 13.825 | 7.877 | 26.046 | 16.409 | 9.648 | 4.983 | 0.500 | 0.286 | 0.034 | 0.000 |
| | sms, message, text, chat | 296.199 | 56.248 | 590.643 | 103.904 | 504.889 | 303.701 | 20.136 | 11.258 | 9.253 | 6.471 |
| | friends, family | 94.296 | 10.529 | 51.458 | 15.938 | 186.250 | 81.009 | 5.812 | 2.842 | 14.014 | 2.792 |
| | voice, voice call, video call, conf call | 25.900 | 9.895 | 5.211 | 1.828 | 6.654 | 1.402 | 6.295 | 1.400 | 6.960 | 1.989 |
| | video, picture | 82.796 | 46.650 | 41.489 | 10.547 | 159.278 | 174.231 | 16.882 | 11.051 | 14.941 | 19.748 |
| | freeze, restart | 1.248 | 9.287 | 0.469 | 6.598 | 28.233 | 208.238 | 0.111 | 2.000 | 3.348 | 33.747 |
| | group, group chat, admin | 20.492 | 10.336 | 13.530 | 2.993 | 71.037 | 40.379 | 0.728 | 1.243 | 0.247 | 0.000 |
| | content, unknow content | 1.583 | 0.411 | 1.389 | 0.000 | 18.479 | 43.233 | 0.000 | 0.000 | 0.108 | 0.418 |
| Delighters | update, beta, beta tester, newer, older | 98.228 | 96.810 | 141.536 | 204.768 | 167.211 | 186.603 | 9.350 | 13.078 | 12.996 | 14.069 |
| | timeline | 0.411 | 0.000 | 53.389 | 32.618 | 0.000 | 0.000 | 9.397 | 14.075 | 0.027 | 0.000 |
| | camera | 0.563 | 0.150 | 1.652 | 0.000 | 28.345 | 24.825 | 0.000 | 0.800 | 11.186 | 12.284 |
| | internet connection | 0.000 | 0.000 | 0.000 | 0.000 | 0.000 | 0.000 | 0.000 | 0.000 | 0.000 | 0.000 |
| | last seen | 0.000 | 0.000 | 0.000 | 0.000 | 0.000 | 0.000 | 0.000 | 0.000 | 0.000 | 0.000 |
| | sticker, emoji | 93.145 | 81.750 | 453.935 | 46.113 | 47.426 | 8.758 | 17.033 | 4.337 | 1.686 | 0.227 |
| | notification | 2.267 | 7.322 | 20.029 | 28.168 | 28.817 | 35.438 | 0.845 | 1.957 | 0.191 | 0.550 |
| | history, chat history | 0.228 | 1.027 | 0.301 | 0.375 | 5.988 | 2.898 | 0.000 | 0.000 | 0.099 | 0.133 |
| Indifferent | offer, spin | 1.541 | 0.000 | 3.031 | 0.000 | 20.829 | 5.186 | 0.000 | 0.000 | 0.097 | 0.052 |
| | voice changer | 0.000 | 0.000 | 0.000 | 0.000 | 0.000 | 0.000 | 0.000 | 0.000 | 0.000 | 0.000 |
| | bubble, chat buble | 0.000 | 0.000 | 1.356 | 0.000 | 27.251 | 7.763 | 0.000 | 0.000 | 0.000 | 0.025 |
| | news, cricket | 2.851 | 0.000 | 67.447 | 8.795 | 0.304 | 3.308 | 1.222 | 4.585 | 1.121 | 1.275 |
| | games, teen patti | 0.000 | 0.000 | 42.323 | 6.337 | 20.773 | 5.330 | 0.400 | 0.000 | 0.252 | 0.082 |
| | profile picture, profile, status, theme, home screen | 18.260 | 6.211 | 77.651 | 30.095 | 64.139 | 72.666 | 5.356 | 6.856 | 0.208 | 0.000 |
| | hashtag | 0.000 | 0.000 | 0.000 | 0.000 | 28.896 | 14.576 | 0.000 | 0.000 | 0.000 | 0.000 |
| Reverse | phone number | 0.000 | 0.000 | 0.000 | 0.000 | 0.000 | 0.000 | 0.000 | 0.000 | 0.000 | 0.000 |

**Table 5:** Manually listed features from app-store description and wiki pages

| Listed Aspects | LINE | WhatsApp | KIK | HIKE | snapchat | Extracted Aspects |
|---|---|---|---|---|---|---|
| Voice Call | Y | Y | | | Y | voice, voice call, |
| Video Call | Y | Y | | Y | Y | video call |
| Group Call | Y | Y | | Y | | group |
| Instant Messaging | Y | Y | Y | Y | Y | message, chat |
| Group chat | Y | Y | Y | Y | | group chat, admin |
| Timeline | Y | | | | | timeline |
| Store File | Y | | | | | |
| Share file | Y | Y | Y | Y | | video, picture |
| Sticker | Y | Y | | Y | | sticker, emoji |
| Official A/c | Paid | | | | | |
| International Calls | Paid | Free | | | | |
| Phone No | | Y | | | | phone number |
| Username | | | Y | | | |
| Offline Messages | | Y | | | | |
| Location | | Y | | | Y | |
| Wall paper | | Y | | | | |
| Notification | | Y | | | | notification |
| Chat history | | Y | | | | history, chat history |
| Message Broadcasting | | Y | | | | |
| Stories | | | | Y | Y | |
| Camera | | | | Y | | camera |
| Photo Editing | | | | Y | | |
| Live Filters | | | | Y | | |
| Hidden Mode | | | | Y | | |
| Share w/o Internet | | | | Y | | |
| Theme | | | | Y | | theme |
| Privacy | | Y | | Y | | end to end encryption, privacy |
| News | | | | Y | | news, cricket |
| Free SMS | | | | Y | | sms, text |
| SD Card Storage | | | | | | |
| Multimedia Chat | | | | | Y | |
| In App Offers / Spins | | | | | | offer, spin |
| In App Game | | | | | | games, teen patti |
| **Recall** | **72.73%** | **66.67%** | **75.00%** | **68.75%** | **50.00%** | **54.55%** |

## 5. Evaluation

To evaluate the extracted features, a list of aspects was gathered form Google app-store description and wiki page of the apps. Then the recall was calculated for each entity separately and as a whole. These results are tabulated in Table 5. According to these measures this simple system performed moderately well. Using the same set of data the overall precision is:

$$precision = \frac{32}{32+19} \times 100 = 62.76\ \%$$

## 6. Future Scope

For this project only simplest methods were used for both aspect extraction and sentiment classification. The pre-processing step did not perfume a spelling correction, which can be incorporated in the future work. The chat like language also creates opportunity for using algorithm like fuzzy matching to find the intended word. Extraction of implicit aspects can also be included in the future work.

The bucketization task was performed by user survey; this part can also be automated using statistical modelling of the reviews along with its original score.

More advanced techniques of sentiment classification can also be employed for more reliable sentiment scores.

## 7. Conclusion

Using simple methods like frequency based aspect extraction and lexicon based sentiment analysis on publicly available data can help product developers to find scope for improvements. A further step of bucketization can help the developers to prioritize these improvement tasks. As it was shown, this method can also be used to compare competitor's products to benchmark a product.